\documentclass[aps,prl,twocolumn,superscriptaddress]{revtex4-2}
\usepackage{epsfig,amsmath,amssymb,amsfonts,graphicx,color,calc,epstopdf}

\usepackage[T1]{fontenc}

\usepackage{xcolor,hyperref}
\hypersetup{
   colorlinks,
   linkcolor={blue!50!black},
   citecolor={blue!50!black},
   urlcolor={blue!80!black}
}

\let\a=\alpha

\let\s=\sigma \let\t=\tau  
   
\let\D=\Delta \let\L=\Lambda

\def\sss{{\bf s}}

\def\MM{{\cal M}}

\def\NN{{\cal N}} 
  
\def\AA{{\cal A}}

\def\ZZ{{\cal Z}}

\def\de{\mathrm{d}}

\newcommand{\beq}{\begin{equation}} 
\newcommand{\eeq}{\end{equation}}
\newcommand{\ba}{\begin{eqnarray}}
\newcommand{\ea}{\end{eqnarray}}

\begin{document}

\title{Stochastic Gradient Descent outperforms Gradient Descent \\
in recovering a high-dimensional signal in a glassy energy landscape}

\author{Persia Jana Kamali}
\affiliation{Universit\'e Paris-Saclay, CNRS, CEA, Institut de physique th\'eorique, 91191, Gif-sur-Yvette, France}

\author{Pierfrancesco Urbani}
\affiliation{Universit\'e Paris-Saclay, CNRS, CEA, Institut de physique th\'eorique, 91191, Gif-sur-Yvette, France}

\begin{abstract}
Stochastic Gradient Descent (SGD) is an out-of-equilibrium algorithm used extensively to train artificial neural networks.
However very little is known on to what extent SGD is crucial for to the success of this technology and, in particular, how much it is effective in optimizing high-dimensional non-convex cost functions as compared to other optimization algorithms such as Gradient Descent (GD).
In this work we leverage dynamical mean field theory to {benchmark} its performances in the high-dimensional limit. {To do that, we} consider the problem of recovering a hidden high-dimensional non-linearly encrypted
signal, a prototype high-dimensional non-convex hard optimization problem. We compare the performances of SGD to GD and we show that SGD largely outperforms GD {for sufficiently small batch sizes}. In particular, a power law fit of the relaxation time of these algorithms shows that the recovery threshold for SGD with small batch size is smaller than the corresponding one of GD.
\end{abstract}

\maketitle

\section*{Introduction}
Stochastic Gradient Descent (SGD) is an optimization algorithm that is at the basis of many deep learning (DL) models trained on large datasets. 
It can be viewed as an approximated Gradient Descent (GD). In GD the gradient of the empirical loss function is used to update the parameters (or weights) of artificial neural networks. It can be usually computed as a sum of terms, each coming from a datapoint in the training dataset. SGD consists in approximating the gradient by summing only datapoints in a particular fraction of the dataset, the minibatch, and then changing the minibatch during time in order to explore the entire dataset (a full exploration of the dataset is called an epoch in training dynamics).
While the main reason for using SGD is clearly the reduction of the computational complexity of the optimization algorithm - computing the full gradient of the loss is costly as soon as the dataset is huge - the effect of SGD on the performances of deep neural networks is largely unknown. Given its ubiquitous use in DL, understanding this point has become a central problem in this research field.

In generic situations, training DL models involves the optimization of a high-dimensional non-convex cost function and a common belief is that SGD introduces a noise that is beneficial for optimization as much as in simulated annealing thermal noise triggers activated process that may lead to barrier jumping and better optimization performances \cite{kirkpatrick1983optimization}. However the extent to which SGD is effective in annealing the empirical loss is largely unknown due to the fact that it is an intrinsically out-of-equilibrium driven dynamics and therefore its stationary measure is unknown. In fact it has been even shown that SGD can easily find bad local minima which perform rather badly \cite{liu2020bad}.

In the last years many research directions have been explored to clarify the effectiveness of SGD.
One of them \cite{neyshabur2014search,soudry2018implicit, gunasekar2018characterizing} has suggested that both GD and SGD dynamics perform an implicit regularization of the neural network parameters during training and this is beneficial for the performances of the networks.
However, especially for the SGD case, these results are often limited to particular optimization settings and therefore this leaves open the possibility that the implicit regularization picture may not hold generically. 
Additionally, the version of SGD considered is typically the one of online (or one-pass) SGD \cite{saad1995line, saad1995exact, saad1997globally, saad1999line, coolen2000dynamics, coolen2000line, veiga2022phase}. In this case, the gradient of the loss is computed via just one datapoint which is never used again. It follows that in this setting there is no notion of training error or epochs and therefore online SGD is slightly far from practical situations. 
{In another research line, it has been suggested that overparametrized models trained via SGD are found in minimizers of the loss which are flat and that flatness positively correlates with generalization performances \cite{yang2021taxonomizing, yang2023stochastic}. However this perspective seems to be challenged by more recent studies \cite{andriushchenko2023modern}. An exact treatment of SGD in these cases is still lacking.}

A parallel research line has focused on understanding the performances of gradient based algorithms in high-dimensional inference problems \cite{mannelli2019passed, mannelli2020marvels, sarao2020complex, sarao2021analytical, mignacco2021stochasticity, ben2022high, angelini2023limits}. {While the loss landscape of these problems is different from the one of overparametrized neural networks where a manifold of minimizers at zero loss is found \cite{draxler2018essentially},} they represent prototypical high-dimensional non-convex hard optimization problems {and, for this reason, they are ideal to benchmark optimization algorithms.} Indeed in these cases, the ground states of the loss function {are sharp minima having} perfect generalization properties but are hard to find given that they are very rare (typically one or two configurations) surrounded by exponentially many spurious minima with bad generalization properties. 
For this class of problems a lot of work has been devoted to study the performances of GD and sampling algorithms such as the Langevin algorithm \cite{mannelli2019passed, mannelli2020marvels, sarao2021analytical} and the performances of online SGD \cite{ben2020algorithmic, arous2021online, ben2022high}.
Recently in \cite{mignacco2021stochasticity} one of these prototypical problems, phase retrieval \cite{chi2019nonconvex, dong2023phase}, has been proposed to asses the effectiveness of the SGD algorithm. 
However, while numerical simulations show that SGD (and in particular a variant of SGD called persistent-SGD) seems to perform better than GD, it is unclear how these results extend in the high dimensional limit. The main difficulty is that the theoretical analysis of SGD is performed using Dynamical Mean Field Theory (DMFT) \cite{cugliandolo2023recent} which however has severe limitations and cannot be used to extrapolate the behavior of the algorithms at long times.

{Our work joins this research line by considering} a different model which has the same landscape structure of previously considered models yet it allows {an exact} theoretical description of the SGD dynamics in the high-dimensional limit. Indeed the DMFT equations are rather different from \cite{mignacco2021stochasticity} and this allows to track SGD dynamics at long times {and small batch size} and to establish quantitatively its superiority with respect to GD.

\section*{The model and the algorithms}
We consider an $N$-dimensional signal $\underline w^*=\{w_1^*,\ldots, w_N^*\}\in {\mathbb R}^N$ uniformly distributed on the sphere $|\underline w^*|^2=N$ and a set of $M=\alpha N$ non-linear measurements
\beq
y_\mu=\frac 1N \sum_{i<j}J^\mu_{ij} w_i^* w_j^*\ \ \ \ \mu=1,\ldots, M
\eeq
where the entries of the symmetric matrices $J^\mu$  are independent and distributed according to a Gaussian law with zero mean and variance given by
\beq
\overline{J^{\mu}_{ij}J^{\nu}_{lm}} = \delta_{\mu\nu}\delta_{il}\delta_{jm}\ \ \ \ i<j\ , \ l<m
\eeq
The control parameter $\alpha$ is called the sample complexity.

A natural inference task is to reconstruct $\underline w^*$ knowing the dataset of measurements and measurement matrices, $\{y_\mu, J^\mu\}_{\mu=1,\ldots, M}$. We expect that this problem becomes easier when $\alpha$ is large and it is more difficult at small sample complexity.
A way to solve it is to define a loss function given by	
\beq
H[\underline x] = \frac 12 \sum_{\mu=1}^M \left(y_\mu-\frac 1N \sum_{i<j}J^\mu_{ij} w_i w_j\right)^2\equiv \sum_{\mu=1}^Mv_\mu(\underline w)\:.
\eeq
The ground state of $H$ has been studied extensively in \cite{fyodorov2019spin} when restricted to the case in which $|\underline w^*|^2$ is fixed.
For $\alpha>1$, $H$ has two trivial ground states given by $\underline w^*$ and $-\underline w^*$ \cite{fyodorov2019spin}. However we will show that the model is glassy and has many spurious minima, poorly correlated with the signal, which do trap the dynamics for sufficiently small $\alpha$ and which coexist with the simple ground state structure.
We are interested in reconstructing $\underline w^*$ by minimizing $H$ via a class of gradient based algorithms
defined by the following dynamical rule
\beq
\underline w(t+1) = \underline w(t) -\frac \eta{b}  \sum_{\mu=1}^M\s_\mu(t) \frac{\partial v_\mu}{\partial \underline w} \:.
\label{GD_based}
\eeq
In this case, time is discretized into small time steps of length $\eta$ also called the learning rate.
The variables $\s_\mu(t)=\{0,\ 1\}$ are selection variables that control which datapoint enters in the computation of the gradient. We take $\s_\mu(t)$ to be random: $\s_\mu(t)=1$ with probability $b\in(0,1]$ and $\s_\mu(t)=0$ with probability $1-b$ \cite{mignacco2020dynamical}. The parameter $b$ controls the batch size, namely the average fraction of datapoints entering the computation of the gradient. We assume that the variables $\s_\mu(t)$ are extracted independently at each time step (even if a time-dependent correlation can be also added in principle \cite{mignacco2020dynamical}) and are totally uncorrelated in the datapoint index $\mu$. The particular case $b=1$ corresponds to GD.  {The initial condition for the dynamics in Eq.~\eqref{GD_based} is assumed to be random with $|\underline w(0)|^2=N$. However \eqref{GD_based} does not constrain the norm of $\underline w$.}

We are interested in measuring the performances of this class of algorithms as a function of the batch size.
This can be done by defining the mean square displacement (MSD) between the signal and the current state of the system $\underline w(t)$
\beq
\Delta(t) = \frac 1N \sum_{i=1}^N |w_i(t)-w_i^*|^2\:.
\eeq
Therefore we would like to study the behavior of $\D(t)$ as a function of the batch size $b$.
Since the SGD noise for $b<1$ is an out-of-equilibrium state-dependent noise, the stationary probability distribution of \eqref{GD_based} is unknown and therefore the only way to extract the performances of the SGD algorithm is to track it directly in the large $N$ limit. This can be done by using dynamical mean field theory (DMFT) \cite{MPV87,cugliandolo2023recent}.

\section*{Dynamical mean field theory}
The MSD can be rewritten as
\beq
\D(t) = 1-2m(t)+C(t,t)
\eeq
where the magnetization $m(t)$ and the correlation function $C(t,t')$ are defined as
\beq
\begin{split}
m(t) &= \underline w(t)\cdot \underline w^*/N \ \ \ \ \ C(t,t') = \underline w(t)\cdot \underline w(t')/N\:.
\end{split}
\label{C_m}
\eeq
Therefore a computation of these dynamical order parameters in the large $N$ limit allows to track the corresponding dynamics of the MSD. 
These quantities concentrate at large $N$ on their typical values. The equations describing their {exact} dynamical behavior can be obtained via a DMFT analysis. For the model described in this work we follow the derivation of the DMFT equations outlined in a related model \cite{kamali2023dynamical, urbani2023continuous, montanari2023solving} and we generalize it to include (i) the fact that the norm of the vector $\underline w(t)$ is unbounded, (ii) the presence of a hidden signal in the loss function and (iii) the effect of the selection variables $\s_\mu(t)$.
Here we present the resulting DMFT equations whose derivation is reported in the Appendix of this manuscript.
\begin{widetext}
{\medmuskip=0mu
\thinmuskip=0mu
\thickmuskip=0mu
These equations are written in terms of an additional order parameter, the response function, given by
\beq
R(t,t') = \lim_{\underline h\to 0}\frac{1}{N}\sum_{i=1}^N \frac{\delta w_i(t)}{\delta h_i(t')}
\eeq
and the field $\underline h(t)$ is introduced in the dynamical equations \eqref{GD_based} via an infinitesimal tilt of the loss function $H\to H-\underline h \cdot \underline w$. Causality of the dynamical equations implies that $R(t,t')=0$ if $t\leq t'$.
The resulting DMFT equations are
\beq
\begin{split}
m(t+1)&= m(t) -\eta^2\a\left(\sum_{s=0}^t\left(\L_R(t,s)C(t,s)+\L_C(t,s)R(t,s)\right)m(s) -m(t)\sum_{s=0}^t\L_R(t,s) \right)\\
C(t+1,t')&=C(t,t')+\eta \Omega_1(t,t') \ \ \ \ \ \ \ \ \forall t'\leq t\\
R(t+1,t') &= \delta_{t,t'}-\eta^2\a \sum_{s=t'+1}^t\left(\L_R(t,s)C(t,s)+\L_C(t,s)R(t,s)\right)R(s,t')\\
C(t+1,t+1) &= C(t,t) + 2\eta \Omega_1(t,t) + \eta^2 \Omega_2(t) \\
\Omega_1(t,t') &=  \a\eta \left[m(t)m(t') \sum_{s=0}^t\L_R(t,s) -\sum_{s=0}^{t'}\L_C(t,s)C(t,s)R(t',s) -\sum_{s=0}^t\left(\L_R(t,s)C(t,s)+\L_C(t,s)R(t,s)\right)C(t',s)\right]\\
\Omega_2(t)&=\a^2\eta^2\sum_{s,s'=0}^t\left(\L_R(t,s)C(t,s)+\L_C(t,s)R(t,s)\right)C(s,s')\left(\L_R(t,s)C(t,s')+\L_C(t,s)R(t,s')\right)\\
&-2\a^2\eta^2 m(t)\left(\sum_{s=0}^t\L_R(t,s)\right)\left(\sum_{s=0}^t\left(\L_R(t,s)C(t,s)+\L_C(t,s)R(t,s)\right)m(s)\right)\\
&+2\a^2\eta^2\sum_{s=0}^t\sum_{s'=0}^s\left(\L_R(t,s)C(t,s)+\L_C(t,s)R(t,s)\right)\L_C(t,s')C(t,s')R(s,s')\\
&-\a \L_C(t,t)C(t,t)+\left(\a\eta\sum_{s=0}^t \L_R(t,s)\right)^2
\end{split}
\label{SP_eqs_unfolded}
\eeq
}
\end{widetext}
and must be integrated starting with an initial condition given by
\beq
m(0)=m_0\ \ \ \ C(0,0)=C_0\:.
\eeq
This means that we are extracting the initial configuration $\underline w(0)$ randomly with a uniform probability distribution such that it has a projection $m_0$ on $\underline w^*$ and its squared norm is $C_0$.  
The 	quantities $\Lambda_C(t,t')$ and $\Lambda_R(t,t')$ play the role of a self-energy in the Dyson equation which describes the evolution of $m$, $C$ and $R$. Their expressions are given by 
\beq
\begin{split}
\Lambda_C(t,t') &=\frac 12 \langle C_A^{(r)}(t,t')\s(t)+C_A^{(l)}(t',t)\s(t') \rangle_\s\\
\Lambda_R(t,t') &=\frac 12 \langle R_A^{(r)}(t,t')\s(t)+R_A^{(l)}(t',t)\s(t') \rangle_\s\
\end{split}
\label{Lambdas}
\eeq
where the random variable $\s(t)$ is a selection variable whose dynamical and statistical properties are the same as a typical selection variable $\s_\mu(t)$. Correspondingly, the brackets denote the average with respect to its probability distribution. Furthermore $C_A^{(l,r)}$ and $R_A^{(l,r)}$ are matrices which depend on $\s(t)$ and whose precise expressions are given in the Appendix.

\section*{Results}
The DMFT equations can be integrated  numerically rather efficiently. At variance with other cases \cite{mignacco2020dynamical, mignacco2021stochasticity} which are typically more difficult, here the only bottleneck is in the computation of the averages appearing in Eq.~\eqref{Lambdas}. This can be done by extracting $\NN$ trajectories of the variable $\s(t)$ and approximating these averages by empirical averages.
In the following we present the results of such integration.
We fix the learning rate $\eta=0.1$ and we take $C_0=1$.
The loss function is invariant under spin-flip and therefore recovering the signal requires a spontaneous breaking of this symmetry. In a finite dimensional system, the small symmetry breaking field is given by the initial condition of the dynamics that always has a finite albeit {of order $N^{-1/2}$} projection on the signal. In order to avoid the inconvenient of analyzing the equation in the small $m_0$ limit we fix its value to be small but finite. This implies that we are considering a hot start for the dynamics. 
Therefore from now on we fix\footnote{{Note that the central limit theorem implies that $m_0=10^{-4}$ is equivalent to having  a system size of order $N\simeq 10^8$ which is impossible to simulate.}} $m_0=10^{-4}$.

\begin{figure}
\includegraphics[width=\columnwidth]{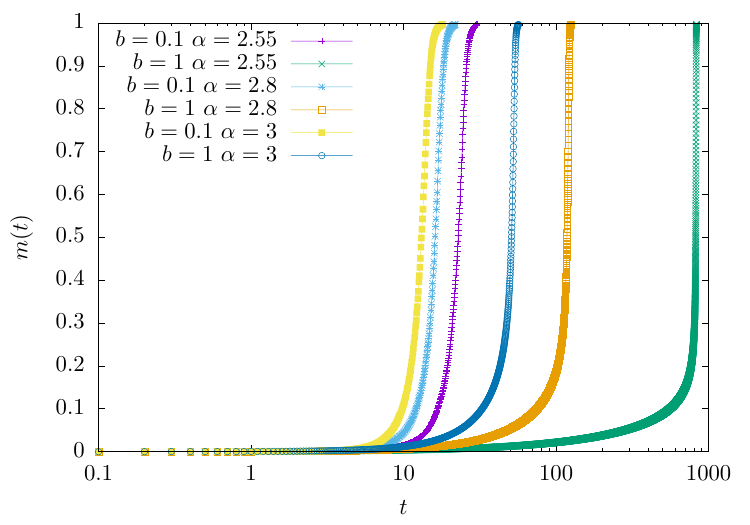}
\caption{The magnetization $m(t)$ as a function of the time $t$ for different values of $\alpha$ and $b$. SGD is systematically faster than GD in reaching the signal. The curves are obtained by integrating the DMFT equations with $\NN=10^3$ samples.}
\label{magnetization_fig}
\end{figure}

\begin{figure}
\includegraphics[width=\columnwidth]{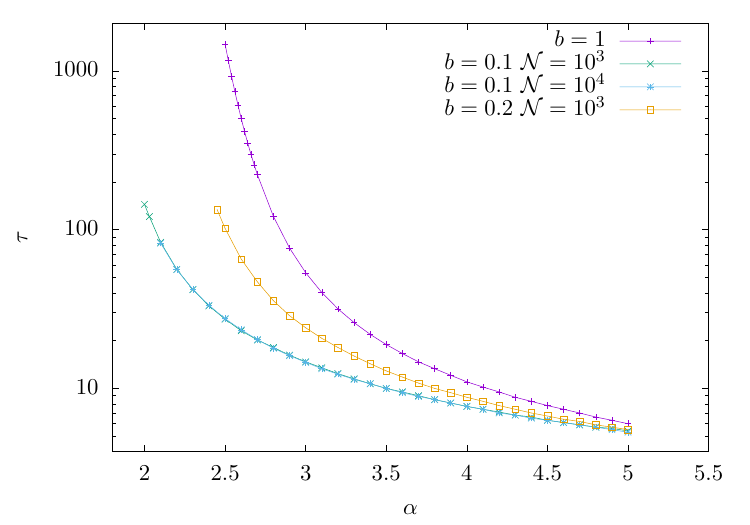}
\caption{The relaxation time to the signal as a function of $\alpha$ for GD and SGD ($b=0.1,0.2$). For $b=0.1$ we show the result of the integration of the DMFT equation for $\s=10^3$ and $\s=10^4$. This shows that for what concerns the relaxation time, with $10^3$ samples we are close to the asymptotic limit.}
\label{relaxation_plot_lin}
\end{figure}

\begin{figure}
\includegraphics[width=\columnwidth]{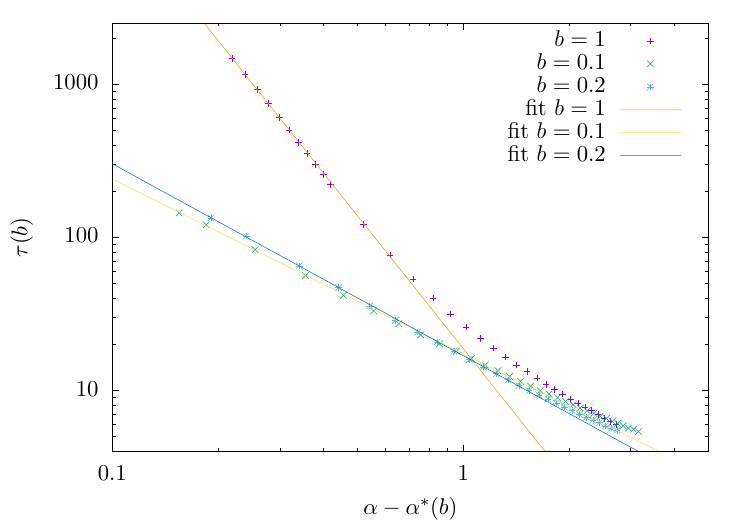}
\caption{Power law fit of the relaxation time with the law in Eq.~\eqref{power_law}. We find that $\alpha^*(b)\simeq \{1.84,\  2.26,\  2.28\}$ for $b=0.1, 0.2, 1$.}
\label{rel_log}
\end{figure}

In order to describe the result of the numerical integration of the DMFT equations we first start by a qualitative analysis. In Fig.\ref{magnetization_fig} we report the behavior of $m(t)$ for GD ($b=1$) and SGD ($b=0.1, 0.2$) and decreasing values of the sample complexity $\a$.
We observe that (i) decreasing $\a$ the time it takes to achieve the signal increases for all algorithms and (ii) the SGD algorithm is qualitatively faster to get to it.
In order to quantify this speedup we define the relaxation time to the signal $\t(b)$ as the time it takes for $\Delta(t)$ to go below a threshold value which we fix to be $0.15$ (and we have checked that shortly after this point all algorithms achieve the signal). In Fig.\ref{relaxation_plot_lin} we plot the behavior of $\t(b)$ for $b=0.1, 0.2, 1$ from which we deduce that the SGD noise provides a clear speedup of the dynamics.
To describe these findings even more quantitatively we assume that the relaxation time diverges as a power law at a recovery transition point $\a^*(b)$ and we fit the data with
\beq
\t(b) \simeq  \t_0(b) |\a - \a^*(b)|^{-z(b)}\:.
\label{power_law}
\eeq
In Fig.\ref{rel_log} we plot the result of this fit.
The fitting value of the recovery threshold $\a^*(b)$ is smaller when $b<1$. In particular for we can compare the cases $b=0.1$ and $b=1$. The recovery threshold for GD is found at $\a^*(1)=2.28$ and for this value of $\a$ SGD with $b=0.1$ has a finite relaxation time to the signal\footnote{{This is the best indication that SGD for $b=0.1$ has a smaller recovery threshold than GD. Indeed fitting the relaxation time in the SGD case is hard given that we have data only on just a decade in time.}}. Therefore the recovery threshold of SGD for small batch size is strictly smaller than the one of GD. This is the main result of our work.

\section*{Discussion}
{To summarize, we have considered a prototypical high-dimensional non-convex hard optimization problem to benchmark the performances of SGD. Thanks to an exact treatment of this algorithm via DMFT, we have shown that SGD at small batch sizes outperforms GD in the sense that the recovery threshold of the former is strictly smaller than the corresponding one of the latter. }

{Close to the recovery threshold, the dynamics is  a two step process: first the system lands on a stationary state not correlated (or poorly correlated) with the signal and then, on longer timescales, this stationary state becomes unstable and the system reaches the ground state of the loss function. This is similar to what happens in the GD case and for the thermal Langevin algorithm \cite{mannelli2019passed, mannelli2020marvels, sarao2020complex}.
However while in these cases the intermediate stationary regime is described by aging \cite{cugliandolo2002dynamics}, in the present case the SGD noise provides an out-of-equilibrium driven dynamics which leads the system away from aging and into a stationary, time-translational invariant (TTI) regime \cite{mignacco2022effective}. Therefore constructing a theory of the recovery threshold for SGD would require a detailed understanding of this intermediate dynamical regime which is left for future work.}

\textbf{Acknowledgments -- } P. J. Kamali acknowledges the Institute of Theoretical Physics of the University Paris-Saclay where this work has been done. P. Urbani wants to thank Antonio Sclocchi for discussions.

%

\appendix
\widetext
\section{Derivation of the DMFT equations}

In this section we derive the dynamical mean field theory equations (DMFT) that describe the behavior of SGD (and GD when $b=1$) in the infinite dimensional limit $N\to \infty$.
The derivation closely follows the one of \cite{kamali2023dynamical} and we will highlight the main differences.

We start with a trivial identity which follows from the causality of the dynamics:
\beq
1 = \mathcal{Z}_{dyn} =\left<\int\mathcal{D}\underline{w}(t)\prod_{i=1}^{N}\delta\left(-\dot{w}_i(t)- \frac{\partial H}{\partial w_i(t)}\right)\right>
\eeq
and the brackets denote an average over the selection variables $\s_\mu$ and the realization of the random matrices $J^\mu$. 
Introducing a Fourier representation for the Dirac delta we get
\beq
\mathcal{Z}_{dyn} = \left<\int\mathcal{D}\underline{w}(t)\mathcal{D} \hat{\underline{w}}(t)\exp\left(i\sum_t\,\hat{\underline{w}}(t)\cdot \left(-\left(\underline w(t+1)+\underline w(t)\right) - \frac \eta b\sum_{\mu}\s_\mu(t)\frac{\partial v_\mu}{\partial \underline{w}(t)}\right)\right)\right>
\eeq
We introduce now Grassmann variables \cite{zinn2021quantum} so that we can rewrite:
\beq
\mathcal{Z}_{dyn} = \left<\int\mathcal{D}\underline{w}(a)\exp\left(-\frac{1}{2}\int da\,db\,\underline{w}(a)\,\mathcal{K}(a,b)\,\underline{w}(b) + M\AA_{loc}\right)\right>
\eeq
and the superfield $\underline w(a)$ is defined on coordinate $a=(t_a,\theta_a)$:
\beq
\underline w(a)=\underline w(t_a) +i\theta_a \underline {\hat w}(t_a) \:.
\eeq
The kernel \(\mathcal{K}(a,b)\) is implicitly defined such that:
\beq
-\frac{1}{2}\int da\,db\,\underline{w}(a)\,\mathcal{K}(a,b)\,\underline{w}(b) = - \sum_t\, i\hat{\underline{w}}(t)\cdot \left(\underline w(t+1)-\underline w(t)\right)
\eeq
Following the same steps as in \cite{kamali2023dynamical} we can show that the term $M\AA_{loc}$ can be obtained as a function of the following dynamical order parameters
\beq
\begin{split}
m(a) &=\frac 1N {\underline{w}^* \cdot \underline{w}(a)}\\
Q(a,b) &= \frac 1N \underline{w}(a) \cdot \underline{w}(b)-m(a)m(b)\:.
\end{split}
\eeq
In particular we have
\begin{equation}
    \begin{split}
        & \mathcal{Z}_{dyn} = \int\mathcal{D}Q\,\mathcal{D}m\,\exp{N\mathcal{A}_{dyn}[Q, m]} \\
        & \mathcal{A}_{dyn} = -\frac{1}{2}\int da\,db\,\mathcal{K}(a,b)\,[Q(a,b)+m(a)m(b)] + \frac{1}{2}\ln\det(Q)+\alpha\AA_{loc}
    \end{split}
\end{equation}
and $\AA_{loc}$ is given by
\begin{equation}
    \begin{split}
        &\AA_{loc} =\frac 12 \ln\left \langle\int\mathcal{D}h(a)\,\mathcal{D}\hat{h}(a)\,e^{S_{loc}^R} \right\rangle+\frac 12 \ln \left \langle \int\mathcal{D}h(a)\,\mathcal{D}\hat{h}(a)\,e^{S_{loc}^L}\right\rangle\\
        &S_{loc}^{(R,L)} = i\int da\, h(a)\hat{h}(a) - \int da\,\sss(a)\frac{h(a)^2}{2} -\frac{1}{2} \int dadb\hat{h}(a)\hat U^{(R,L)}(a,b)\hat{h}(b)
    \end{split}
    \label{effective_proc}
\end{equation}
and we have defined $U^{(R,L)}$ as
\beq
\begin{split}
\hat U^{(L)}(a,b) &= \sss(a)\left[\frac{1}{2} - \frac{m(a)^2}{2} -\frac{m(b)^2}{2} + \frac{(Q(a,b) + m(a)m(b))^2}{2}\right]\\
\hat U^{(R)}(a,b) &= \left[\frac{1}{2} - \frac{m(a)^2}{2} -\frac{m(b)^2}{2} + \frac{(Q(a,b) + m(a)m(b))^2}{2}\right]\sss(b)
\end{split}
\eeq
and $\sss(a)$ is an effective selection variable which has only a scalar component:
\beq
\sss(a) = \frac{\s(t)}b\:.
\eeq
The brackets in Eq.~\eqref{effective_proc} denote the average with respect to $\sss$.
Performing the Gaussian integral over $\hat{h}$ and $h$ we find
\begin{equation*}
    \begin{split}
        &\AA_{loc}\propto \frac 12 \ln\langle \det(I + \hat U^{(R)})^{-\frac{1}{2}}\rangle  + \frac 12 \ln \langle \det(I + \hat U^{(L)})^{-\frac{1}{2}} \rangle
    \end{split}
\end{equation*}
and we have neglected irrelevant constant factors.
The integral defining $\ZZ_{dyn}$ can the be evaluated through a saddle point. 
Taking the functional derivatives with respect to $Q(a,b)$ and $m(a)$ and then performing the shift $Q(a,b)\to Q(a,b)-m(a)m(b)$ we get the following equations: 
\begin{equation}
    \begin{split}
        &0=- \int \de c\,\mathcal{K}(a,c)Q(c,b) + \delta(a,b) -\alpha \int \de c\,\Lambda(a,c)Q(a,c)Q(c,b) - m(a)m(b)\int \de c \Lambda(a,c)
    \end{split}
\end{equation}
\begin{equation}
    \begin{split}
        0=- \int \de b \mathcal{K}(a,b)m(b) - \alpha\left[ \int\de c \Lambda(a,c)Q(a,c)m(c)-m(a)\int\de c\Lambda(a,c)\right] 
    \end{split} 
    \label{sp_eqs}
\end{equation}
and $\L(a,b)$ is given by
\beq
\Lambda(a,b)= \frac 12 \left\langle \left(1+U^{(L)}\right)^{-1}(a,b)\sss(b) +\left(1+U^{(R)}\right)^{-1}(a,b)\sss(a) \right\rangle
\label{def_Lambda_folded}
\eeq
where
\beq
\begin{split}
U^{(L)}(a,b) &= \sss(a)\left[\frac{1}{2} - \frac{m(a)^2}{2} -\frac{m(b)^2}{2} + \frac{Q(a,b)^2}{2}\right]\\
U^{(R)}(a,b) &= \left[\frac{1}{2} - \frac{m(a)^2}{2} -\frac{m(b)^2}{2} + \frac{Q(a,b)^2}{2}\right]\sss(b)\:.
\end{split}
\eeq
At this point we can unfold the Grassmann structure of the equations. In order to do that one can show  by inspection that the Grassmann structure of the fields at their saddle point value is given by 
\begin{equation}
    \begin{split}
        &Q(a,b) = C(t_a,t_b)+ \theta_aR(t_b,t_a) + \theta_bR(t_a, t_b) \\
        &\L (a,b) = \L_C(t_a,t_b) + \theta_a\L_R(t_b,t_a) + \theta_b\L_R(t_a, t_b)\\
        &m(a) = m(t_a)
    \end{split}
\end{equation}
Substituting these expressions inside Eqs.\eqref{sp_eqs} we get Eqs.~\eqref{SP_eqs_unfolded} of the main text.
We note that the Eqs.\eqref{sp_eqs} do not contain the equation to propagate the diagonal part of the matrix $C(t,t)$. This can be derived either by taking the corresponding derivatives of $\AA_{dyn}$ with respect to these quantities or by following an alternative route which we explain now given that it is easier.
The main idea is to write a self consistent process for an effective variable $w$. Using the same strategy as in \cite{castellani2005spin}, we get to
\beq
w(t+1) =w(t) -\alpha\eta^2 \sum_{s=0}^t \left[\L_R(t,s)C(t,s)+\L_C(t,s)R(t,s)\right]w(s) + \alpha \eta^2 m(t) \sum_0^t\L_R(t,s)+\xi(t) 
\eeq
where the noise $\xi$ has zero average and two point correlation function given by
\beq
\langle\xi(t)\xi(s)\rangle =- \alpha \L_C(t,s)C(t,s)\:.
\eeq
Therefore we get that
\beq
C(t+1,t+1)= \left\langle \left(w(t) -\alpha\eta^2 \sum_{s=0}^t \left[\L_R(t,s)C(t,s)+\L_C(t,s)R(t,s)\right]w(s) + \alpha \eta^2 m(t) \sum_0^t\L_R(t,s)+\xi(t)\right)^2\right\rangle_\xi
\eeq
which gives the corresponding equation reported in Eq.~\eqref{SP_eqs_unfolded}.
In order to conclude our derivation we need to give precise expressions for the kernels $\L_C(t,t')$ and $\L_R(t,t')$. These can be derived unfolding the Grassmann structure in their definition in Eq.~\eqref{def_Lambda_folded}. In particular we obtain Eqs.~\eqref{Lambdas} of the main text where $C_A^{(l)}$, $C_A^{(r)}$, $R_A^{(l)}$ and $R_A^{(r)}$ solve the following equations
\begin{equation}
\begin{split}
&\sum_{s=0}^t \MM^{(R)}(t',s)         \begin{pmatrix}
            C_A^{(r)}(t,s)\\
            R_A^{(r)}(t,s)
        \end{pmatrix}
        = 
        \begin{pmatrix}
            0\\
            \delta_{t,t'}/\eta
        \end{pmatrix}\ \ \ \ \ t'\leq t\\
& \MM^{(R)}(t',s) =\begin{pmatrix}
            \delta_{s,t'} + \frac \eta b C(s,t')\,R(t',s)\s(t') & \frac{\eta}{2 b}[1-m(s)^2-m(t')^2 + C(s,t')^2]\s(t')\\
            0 & \delta_{s,t'} + \frac \eta b C(s,t')\,R(s,t')\s(t')
        \end{pmatrix}\:.
\end{split}
\end{equation}
\begin{equation}
\begin{split}
&\sum_{s=0}^t \MM^{(L)}(t',s)         \begin{pmatrix}
            C_A^{(l)}(t,s)\\
            R_A^{(l)}(t,s)
        \end{pmatrix}
        = 
        \begin{pmatrix}
            0\\
            \delta_{t,t'}/\eta
        \end{pmatrix}\ \ \ \ \ t'\leq t\\
& \MM^{(L)}(t',s) =\begin{pmatrix}
            \delta_{s,t'} + \frac \eta b C(s,t')\,R(t',s)\s(s) & \frac{\eta}{2 b}[1-m(s)^2-m(t')^2 + C(s,t')^2]\s(s)\\
            0 & \delta_{s,t'} + \frac \eta b C(s,t')\,R(s,t')\s(s)
        \end{pmatrix}\:.
\end{split}
\end{equation}
This concludes the derivation of the DMFT equations. We underline that they have a causal structure and therefore they can be easily integrated numerically.
We underline that the numerical integration for the GD case, $b=1$, is computationally simpler given that there is no need to compute the averages of Eqs.~\eqref{Lambdas} of the main text. This is the reason why we systematically achieve longer times in the GD case.

\end{document}